\documentclass[letterpaper, 10 pt, conference]{ieeeconf}

\IEEEoverridecommandlockouts                              

\usepackage{amssymb, amsmath, mathtools}
\usepackage{cite}
\usepackage[breaklinks=true,colorlinks,citecolor=blue,linkcolor=blue,urlcolor=blue]{hyperref}
\usepackage{soul}
\usepackage{comment}
\usepackage{tabularx, booktabs}
\usepackage{xcolor}
\usepackage{multirow}
\usepackage{subfigure}
\newcommand{\figlabel}[1]{Figure~\ref{#1}}
\newcommand{\tablabel}[1]{Table~\ref{#1}}
\newcommand{\seclabel}[1]{Section~\ref{#1}}

\newcommand\crule[1][black]{\textcolor{#1}{\rule{0.2cm}{0.2cm}}}

\newcolumntype{Y}{>{\centering\arraybackslash}X}

\definecolor{Wine}{RGB}{150,0,90}
\definecolor{DarkBlue}{RGB}{17,0,188}
\definecolor{Blue}{RGB}{0,19,243}
\definecolor{LightBlue}{RGB}{0,110,255}
\definecolor{Teal}{RGB}{24,209,197}
\definecolor{Green}{RGB}{92,255,83}
\definecolor{Lime}{RGB}{186,248,0}
\definecolor{Yellow}{RGB}{255,207,0}
\definecolor{Orange}{RGB}{255,99,0}
\definecolor{Red}{RGB}{255,0,0}

\title{\LARGE \bf
Solving Logistic-Oriented Bin Packing Problems Through a Hybrid Quantum-Classical Approach
}

\author{Sebastián V. Romero$^{1}$, Eneko Osaba$^{1\dagger}$, Esther Villar-Rodriguez$^{1}$ and Antón Asla$^2$
\thanks{This work was supported by the Basque Government through HAZITEK program (Q4\_Real project, ZE-2022/00033), and by the Spanish CDTI through Proyectos I+D Cervera 2021 Program (QOptimiza project, 095359).}%
\thanks{$^{\dagger}$Corresponding Author.}
\thanks{$^{1}$Sebastián V. Romero, Eneko Osaba, and Esther Villar-Rodriguez are with TECNALIA, Basque Research and Technology Alliance (BRTA), 48160 Derio, Spain. Contact email: eneko.osaba@tecnalia.com.}
\thanks{$^{2}$Antón Asla is in Serikat - Consultoría y Servicios Tecnológicos, 48009 Bilbao, Spain.}%
\thanks{© 2023 IEEE. Personal use of this material is permitted. Permission from IEEE must be obtained for all other uses, in any current or future media, including reprinting/republishing this material for advertising or promotional purposes, creating new	collective works, for resale or redistribution to servers or lists, or reuse of any copyrighted	component of this work in other works.}
}

\begin{document}

\maketitle

\begin{abstract}
The Bin Packing Problem is a classic problem with wide industrial applicability. In fact, the efficient packing of items into bins is one of the toughest challenges in many logistic corporations and is a critical issue for reducing storage costs or improving vehicle space allocation. In this work, we resort to our previously published quantum-classical framework known as \texttt{Q4RealBPP}, and elaborate on the solving of real-world oriented instances of the Bin Packing Problem. With this purpose, this paper gravitates on the following characteristics: \textit{i)} the existence of heterogeneous bins, \textit{ii)} the extension of the framework to solve not only three-dimensional, but also one- and two-dimensional instances of the problem, \textit{iii)} requirements for item-bin associations, and \textit{iv)} delivery priorities. All these features have been tested in this paper, as well as the ability of \texttt{Q4RealBPP} to solve real-world oriented instances.
\end{abstract}%

\section{Introduction}

The Bin Packing Problem (BPP) is a combinatorial optimization problem that consists of packing a group of items, or packages, into a set of bins or containers without any kind of overlap. Usually, the main objective of the BPP is \textit{i}) to store all items in a single bin as densely as possible or \textit{ii}) to distribute the complete set of packages into as few containers as possible.

The BPP is a classical problem in artificial intelligence and operations research, and it has wide industrial applicability. The BPP has been applied to problems ranging from origami design \cite{bruton2016packing}, steel production \cite{maddaloni2016bin}, or placement of advertisements \cite{kim2020online}, among many others. Furthermore, transportation and logistics are application fields that have benefited from the advancements made regarding the BPP.

In fact, one of the toughest challenges for logistic corporations is related to bin packing, which is a critical issue for reducing storage costs or improving vehicle space allocation \cite{tian2023learning}. Among general applications, we can highlight pallet loading \cite{gzara2020pallet}, automatic depot management \cite{you2014design}, or optimizing shared parking resources \cite{zhao2021mathematical} to name a few examples.

Due to their remarkable scientific and business interest, BPP variants have been widely addressed in the literature by different solving approaches. In this paper, we explore the BPP from a perspective still scarcely studied in the literature: Quantum Computing (QC, \cite{gyongyosi2019survey}). Nowadays, QC is attracting a lot of attention from researchers and practitioners, and the related community is working to discern the contributions that this revolutionary technology can offer to many application sectors. Transportation and logistics are fields in which QC has already demonstrated promising performance \cite{wang2021shaping}, for instance, 
tackling problems related to routing \cite{osaba2022systematic} or traffic management \cite{boyapati2020quantum}. 
In spite of this blossoming scientific activity, BPP is still 
few dealt from QC perspective.

Thus, our main objective is to take a step forward with respect to our previous study presented in \cite{tecnalia2023hybrid}. In that work, we proposed a QC-based framework for dealing with real-word three dimensional BPPs (3dBPP, \cite{martello2000three}). We coined this framework \textit{Quantum Framework for Real Bin Packing Problems} (\texttt{Q4RealBPP}). In this manuscript, we further elaborate on the formulation and findings presented in \cite{tecnalia2023hybrid} by considering the following aspects:
\begin{itemize}
    \item 
    Contemplate having heterogeneous bins in the same instance, with different sizes and capacities.
    \item 
    Extend the applicability of \texttt{Q4RealBPP} for solving both one-dimensional (1dBPP, \cite{munien2021metaheuristic}) and two-dimensional (2dBPP, \cite{lodi2002recent}) instances. To the best of our knowledge, this feature becomes \texttt{Q4RealBPP} in the first QC-based framework in the literature for dealing with not only real-world oriented 3dBPP problems but also with 2dBPP instances.
    \item 
    Allow the user to impose requirements for item-bin associations. In other words, each item must be assigned to a specific set of bins.
    \item 
    Embrace the relative positioning feature introduced in \cite{tecnalia2023hybrid} (and later detailed), 
    exploring the application of delivery priorities 
    to store items. This way, the user defines which packages will be delivered first, so that they are placed closer to the trunk's door or bin boundaries.
\end{itemize}

All these new features enable to outline more realistic BPP instances with a more accurate applicability to the field of logistics. Therefore, in this paper, we describe how these features have been implemented into \texttt{Q4RealBPP}. Also, we conduct experiments for solving real-oriented instances over new scenarios not contemplated at all in the original formulation, proving the flexibility of our methodology.

It is interesting to mention that all the advances detailed in this research have been defined involving a national company specialized in transport and logistics, named Ertransit\footnote{https://www.ertransit.com.cn/ertransit-espana/}. After conducting several meetings among Ertransit and the research team, a set of real needs were established, which have guided the developments detailed in this paper.

The rest of the article is organized as follows: in Section \ref{sec:back}, we provide the background of this research. Then, in Section \ref{sec:Q4RealBPP} we describe \texttt{Q4RealBPP} and all the newly implemented aspects. Section \ref{sec:demo} is devoted to demonstrating the applicability of the new version of \texttt{Q4RealBPP}. Finally, we conclude this article by summarizing the work done and proposing possible future research in~\seclabel{sec:conc}.

\section{Background}\label{sec:back}

The BPP, in any of its variants, has been extensively studied over the last decades. In fact, interest is so strong that significant research is still being done on even the most basic version, the 1dBPP \cite{munien2021metaheuristic}. Furthermore, the 2dBPP is also widely studied, having practical applications in the sheet metal, leather, clothing, and glass industries \cite{hu2020greedy}. In these cases, the 2dBPP is employed to lay out some pieces on a surface minimizing waste of material. The 2dBPP has also demonstrated applicability in business transportation, giving rise to the well-known \textit{vehicle routing problem with two-dimensional loading constraints} (2L-CVRP, \cite{leung2011extended}).



So far, the 3dBPP is the BPP version that has been explored the most. Highlighted in studies such as \cite{lodi2002heuristic}, the 3dBPP has relevance in many industrial settings. Also, thanks to its complexity, the 3dBPP is also employed as a benchmarking problem to test new techniques and procedures \cite{paquay2018mip}. In any case, it is hard to find 3dBPP instances that can really handle the size and requirements expected by industry \cite{gzara2020pallet}. For this reason, even though the most basic formulation does not faithfully respond to the demands of real-world scenarios, the 3dBPP is also used as a subproblem of more complex and larger problems \cite{parreno2010hybrid}. Additionally, a wide variety of practical constraints have been included to model more realistic 3dBPP variants~\cite{do2021practical}, such as load balancing, uncertainty, overweight restrictions, priority, or load bearing, among many others. 

Regarding its solving, BPP-related variants have been approached using different techniques, such as exact methods \cite{silva2019exact}, classical local-search heuristics \cite{jiang2012hybrid}, nature-inspired metaheuristics \cite{weiss2021solving} and more advanced paradigms, such as reinforcement learning \cite{hu2017solving}. Anyway, if we pay attention to the QC field, BPPs have drawn scant attention so far.


In a nutshell, QC is a revolutionary computing paradigm that leverages quantum phenomena to find competent solutions in a more efficient way. For doing that, specific processing units called Quantum Processing Units (QPU) are used. Up until now, the potential of QC has been explored in research areas such as simulation, machine learning, cryptography, or optimization. The rapid evolution of related technology and the advances made in its democratization \cite{seskir2023democratization} have contributed to the growth of QC community, especially in the optimization field, in which areas as transportation \cite{harwood2021formulating}, energy \cite{ajagekar2019quantum} or finance \cite{orus2019quantum} have already benefited from this avant-garde computing paradigm.

At the confluence between QC and BPP, as mentioned, only a few dedicated papers can be found in the literature to date. It was not until 2022 that the pioneering works focused on 1dBPP were published \cite{de2022hybrid,garcia2022comparative}. Besides those studies, in \cite{bozhedarov2023quantum} the problem of filling spent nuclear fuel in deep-repository canisters is addressed by a 1dBPP approach. Authors solve the problem using the D-Wave quantum annealer. 

Lastly, the authors of this paper presented a framework for solving real-world oriented 3dBPP instances in 2023 \cite{tecnalia2023hybrid}. This framework, named \texttt{Q4RealBPP}, and built upon an existing code \cite{3dBPP}, allows the consideration of features such as \textit{i}) dimensions of items and containers, \textit{ii}) overweight restrictions, \textit{iii}) affinities among items, \textit{iv}) relative positioning of items and \textit{v}) load balancing. In addition, the authors developed a Python script for the generation of synthetic datasets. We refer interested readers to \cite{osaba2023benchmark}.

As mentioned before, this research improves significantly this previous work adding further features and novel procedures to deal with even more realistic and varying scenarios.

\section{Quantum Framework for Real Bin Packing Problems - \texttt{Q4RealBPP}}\label{sec:Q4RealBPP}

\texttt{Q4RealBPP} is a quantum-based framework making use of Leap Constrained Quadratic Model (CQM) Hybrid Solver~(\texttt{LeapCQMHybrid}, \cite{leapCQM}) as its quantum engine. As mentioned, five different features were taken into account in \texttt{Q4RealBPP}: \textit{i}) dimensions, \textit{ii}) maximum weight allowed per bin, \textit{iii}) affinities among packages, \textit{iv}) relative positioning of items, and \textit{v}) the establishment of a center of mass. Lastly, we finish this section by implementing an additional mechanism with the aim of optimizing the inner space between items for reaching better results.

In this article, we add a new functionality layer on top of \texttt{Q4RealBPP}, seeking to tackle a broader spectrum of realistic scenarios, especially focused on logistics environments. This section describes the new contributions. The parameters and variables used in our formulation are shown in~\tablabel{tab:params_vars_used}.
\noindent\begin{table*}[!t]
\centering
\caption{Parameters and variables used in this manuscript.}\label{tab:params_vars_used}\vspace{-2mm}%
\begin{tabularx}{\linewidth}{lX}
\toprule
\multicolumn{1}{l}{\bf Parameters} \\
$I,J,K_i,Q$ & sets of items, bins, orientations of $i$-th item and relative positions between items. \\
$m,n$ & number of items and bins. \\
$l_i,w_i,h_i,\mu_i$ & length, width, height and weight of item $i\in I$. \\
$L_j,W_j,H_j,M_j$ & length, width, height and maximum capacity (\textit{optional}) of bin $j\in J$. \\
$P^+_q$ & set of relative positionings $q\in Q$ between pairs of items $i,k\in I$ to satisfy (\textit{optional}). \\
$(\tilde{L},\tilde{W})$ & target center of mass of the resultant packings (\textit{optional}). \\ [1mm]
\multicolumn{1}{l}{\bf Variables} \\
$v_j$ & binary variable that represents if bin $j\in J$ is used. \\
$u_{i,j}$ & binary variable that represents if item $i$ is added to bin $j$. \\
$r_{i,k}$ & binary variable that shows if orientation $k\in K_i$ is applied to item $i$. Used to compute effective length, width and height $(x'_i,y'_i,z'_i)$. \\
$x_i,y_i,z_i$ & real variables that return the location of the back lower left corner of item $i$ along $x$, $y$ and $z$ axes. \\
$b_{i,k,q}$ & binary variable that returns the relative position $q\in Q$ between items $i,k\in I$. \\ \bottomrule
\end{tabularx}
\end{table*}%

\textbf{Heterogeneous bins:} as can be read in papers such as \cite{martello2000three,lodi2002recent}, canonical BPPs count with a set of unlimited bins with the same size and capacity. Arguably, this is the most frequent trend in the literature, which is far away from real-world settings. This is particularly acute in road transport, where vehicle fleets are usually heterogeneous. In our previous formulation, every bin has the same dimensions~\cite{tecnalia2023hybrid}. Here, we extend its capabilities by considering $n$ bins of heterogeneous dimensions $\{(L_j,W_j,H_j)\}_{j=1}^n$. This feature is added within bin boundary constraints $\forall i\in I,\text{ }\forall j\in J$ as
\begin{align}\label{eq:x_in_bins}
 &x_i+x_i'-\sum\nolimits_{p=1}^jL_p \le (1-u_{i,j})\sum\nolimits_{p=1}^n L_p, \\ \label{eq:x_in_bin_j}
 &x_i-u_{i,j}\sum\nolimits_{p=1}^{j-1}L_p \ge 0\quad(j>1), \\ \label{eq:y_in_bin_j}
 &y_i+y_i'-W_j \le (1-u_{i,j})W_\text{max}, \\ \label{eq:z_in_bin_j}
 &z_i+z_i'-H_j \le (1-u_{i,j})H_\text{max}.
\end{align}
with $W_\text{max}=\max\{W_i\}_{i=1}^n$ and $H_\text{max}=\max\{H_i\}_{i=1}^n$.
For 2dBPP, just consider~\eqref{eq:x_in_bins}-\eqref{eq:y_in_bin_j} and for 1dBPP~\eqref{eq:x_in_bins} and~\eqref{eq:x_in_bin_j}.

Similarly, instead of only having the same maximum capacity for all bins, it is possible to set one for each bin as $\{M_j\}_{j=1}^n$ and modify the overweight restriction as
\begin{equation}\label{eq:max_cap}
 \sum\nolimits_{i=1}^m \mu_iu_{i,j}\le M_j\quad\forall j\in J.
\end{equation}

\textbf{1dBPP/2dBPP compatibility:} being the most complete variant of the BPP family, the 3dBPP is more conducive to solving transportation-related problems. However, 2dBPP and 1dBPP have also proven to be effective formulations for modeling real-world scenarios. The former is a key component of the aforementioned 2L-CVRP, while the 1dBPP has also been embraced for routing environments~\cite{anand2020bin}. With the principal motivation of extending the applicability of the framework, we have generalized \texttt{Q4RealBPP} to cover 1dBPP and 2dBPP problems, letting the user codify the problem in as many spatial dimensions as the problem demands. This new functionality has augmented the capabilities of the framework presented in \cite{tecnalia2023hybrid}.

For 1dBPP, where items are represented just by their $l_i$, orientations are ignored. The effective lengths are $x_i'=l_i$ $\forall i\in I$. This is not the case of 2dBPP, in which items are described by their $l_i$ and $w_i$, and where one rotation can be applied to each of them. The set of non-redundant orientations $K_i$ for each item $i$ is $\varnothing$ if $l_i=w_i$ (square items) and $\{1,3\}$ otherwise (see~\figlabel{fig:orientations}), needing $\kappa=\sum_{i=1}^m|K_i|$ orientation variables $r_{i,k}$. For $i\in I_\text{s}$ with $I_\text{s}=\{i\in I\,|\, l_i=w_i\}$, we can set in advance $r_{i,1}=1$ and $r_{i,3}=0$. Their effective lengths and widths are $x_i'=l_ir_{i,1}+w_ir_{i,3}$ and $y_i'=w_ir_{i,1}+l_ir_{i,3}$ $\forall i\in I$. 

Since the dimensionality of the problem changes, some restrictions of~\cite{tecnalia2023hybrid} must be accordingly modified. For 2dBPP, non-overlapping constraints are given by $\forall i,k\in I,\,\forall j\in J$,
\begin{align}\label{eq:geo0_2dbpp}
    &(u_{i,j}u_{k,j}+b_{i,k,1}-2)\!\sum\nolimits_{p=1}^n L_p+x_i+x'_i-x_k\le0, \\ \label{eq:geo1_2dbpp}
    &(u_{i,j}u_{k,j}+b_{i,k,2}-2)W_\text{max}+y_i+y'_i-y_k\le0, \\ \label{eq:geo3_2dbpp}
    &(u_{i,j}u_{k,j}+b_{i,k,4}-2)\!\sum\nolimits_{p=1}^n L_p+x_k+x'_k-x_i\le0,\\ \label{eq:geo4_2dbpp}
    &(u_{i,j}u_{k,j}+b_{i,k,5}-2)W_\text{max}+y_k+y'_k-y_i\le0,
\end{align}%
where~\eqref{eq:geo0_2dbpp} considers item $i$ at the left of item $k$ ($q=1$),~\eqref{eq:geo1_2dbpp} that $i$ is behind $k$ ($q=2$),~\eqref{eq:geo3_2dbpp} that $i$ is at the right of $k$ ($q=4$) and~\eqref{eq:geo4_2dbpp} that $i$ is in front of $k$ ($q=5$). Similarly, for 1dBPP, we just need to consider~\eqref{eq:geo0_2dbpp} and~\eqref{eq:geo3_2dbpp}. \figlabel{fig:positions} depicts the set of relative positions among items.%
\noindent\begin{figure}[!b]
 \vspace{-5mm}
 \centering
 \subfigure[$k = 1$]{\includegraphics{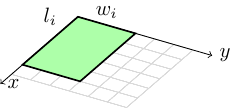}}%
 \subfigure[$k = 3$]{\includegraphics{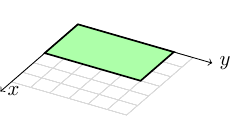}}\vspace{-2mm}%
 \caption{Orientations $k\in K_i$ for an item $i$ of dimensions $(l_i,w_i)$.}\label{fig:orientations}
\end{figure}%
\noindent\begin{figure*}[!t]
    \centering
    \subfigure[$b_{\textcolor{red}{i},\textcolor{blue}{k},1}=1$]{\includegraphics{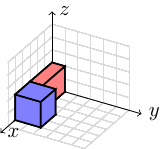}}\hspace{1mm}%
    \subfigure[$b_{\textcolor{red}{i},\textcolor{blue}{k},2}=1$]{\includegraphics{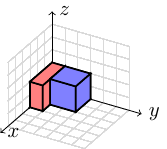}}\hspace{1mm}%
    \subfigure[$b_{\textcolor{red}{i},\textcolor{blue}{k},3}=1$]{\includegraphics{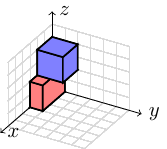}}\hspace{1mm}%
    \subfigure[$b_{\textcolor{red}{i},\textcolor{blue}{k},4}=1$]{\includegraphics{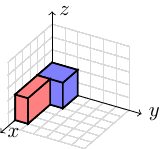}}\hspace{1mm}%
    \subfigure[$b_{\textcolor{red}{i},\textcolor{blue}{k},5}=1$]{\includegraphics{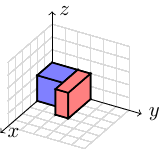}}\hspace{1mm}%
    \subfigure[$b_{\textcolor{red}{i},\textcolor{blue}{k},6}=1$]{\includegraphics{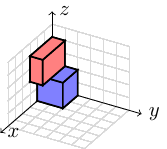}}\vspace{-2mm}%
    \caption{Scheme of $b_{i,k,q}$ activated for relative positions $q\in Q$ among items $i,k\in I$. Both are in contact but it is not mandatory.}\label{fig:positions}
\end{figure*}%

\textbf{Requirements for item-bin association:} with this new functionality, the user is able to assign a set of eligible bins for each item. This way, the solution is considered feasible if all items are allocated to an admissible container. This feature is especially interesting in environments in which bins are heterogeneous and of different characteristics. We can easily think of a scenario with a mixture of refrigerated and non-refrigerated containers. In these cases, it is common to have certain products that need to be kept cold, whereas others may not tolerate low temperatures.

This new feature adds for each item category $\alpha$ a list, $J_\alpha$, of available bins as input, i.e. a set of bins where $\alpha$-items must be placed. Defining $I_\alpha=\{i\in I\,|\,\texttt{id}\text{ of }i\text{ is }\alpha\}$, there are three possible scenarios depending on the length of $J_\alpha$:
\begin{enumerate}
 \item $|J_\alpha|=1$: then $\alpha$-items must be placed in only one bin. Thus, we can set in advance $u_{i,j}=1$ $\forall i\in I_\alpha,\text{ }j\in J_\alpha$ and $u_{i,j}=0$  $\forall i\in I_\alpha,\text{ }\forall j\notin J_\alpha$.
 \item $1<|J_\alpha|<n$: then $\alpha$-items must be placed in certain bins, so some of them are excluded. So, we can set in advance $u_{i,j}=0$  $\forall i\in I_\alpha,\text{ }\forall j\notin J_\alpha$.
 \item $|J_\alpha|=n$: $\alpha$-items can be placed in any bin. That is, no requirement for item-bin association is applied. Hence, we follow the same procedure as in~\cite{tecnalia2023hybrid}.
\end{enumerate}
Taking this into account, $\sum_\alpha (n-|J_\alpha|)|I_\alpha|$ variables can be set in advance, leading to a formulation where fewer variables (thus qubits) are needed (suitable for NISQ era~\cite{Preskill2018}). 

\textbf{Delivery priorities:} in academia, BPP and routing problems are usually treated separately. In contrast, in a real logistics environment, these two problems are interrelated in such a way that the global solution satisfies all the requirements coherently. Thus, optimality in terms of time would involve taking into account the delivery ordering of the customers assigned to each vehicle in such a way that items to be delivered first are placed closest to the unloading door. It is easy to imagine a scenario in which the items to be delivered first are placed closest to the bin boundaries.

We have materialized these scenarios by using the functionality, described in \cite{tecnalia2023hybrid}, named \textit{relative positioning among items}. In our previous work, we employed it for load-bearing purposes, considering the weight of items as a reference for fragility. In that case, heavy
packages should be placed beneath the lightest ones, eventually controlling $b_{i,k,q}$ variables, specifically in this example with $q=3$ and $q=6$ (see~\figlabel{fig:positions}). Therefore, similarly to~\cite{tecnalia2023hybrid}, delivery priorities are specified to influence the values to some $b_{i,k,q}$ variables, 
we define delivery priorities as:
\begin{equation}
\begin{aligned}
    P^+_q&=\{(i,k)\in I^2\,|\,i<k\text{ and }i\text{ delivered before }k\}, \\
    P^+_{q'}&=\{(i,k)\in I^2\,|\,i<k\text{ and }k\text{ delivered before }i\}, 
\end{aligned}
\end{equation}
being $q=1$ and $q'=4$ in the case of 3dBPP. So, we set $b_{i,k,1}=1$ and $0$ otherwise for $(i,k)\in P^+_1$ and $b_{i,k,4}=1$ and $0$ otherwise for $(i,k)\in P^+_4$. For 2dBPP, the same procedure is followed, but using $q=2$ and $q'=5$.

Note that this constraint, and all of those relating relative positions, could be easily adapted to any other scenario via $b_{i,k,q}$. For example, to problems with bins that are accessed from the top ($z$ axis) or from their laterals ($y$ axis), such as with tarpaulin trailers. 

\textbf{Reduce inner space between items:} 2dBPP has well-known significance where cutting rectangular items from raw material sheets, such as steel or wood, is required minimizing the amount of left-overs~\cite{cid2020positions}. Also, for the 3dBPP, a user may prefer to have all items stacked, with the intention of reducing the space between packages. For materializing this requirement, one key constraint consists on pushing items to boundaries, which can be introduced using
\begin{equation}
 o_x = \frac{1}{mL}\sum\nolimits_{i=1}^m(x_i+x_i')\,\text{ } o_y = \frac{1}{mW}\sum\nolimits_{i=1}^m(y_i+y_i'),
\end{equation}
which extends the second objective of~\cite{tecnalia2023hybrid}.


\section{Demonstration}\label{sec:demo}

In this section, we demonstrate the applicability of our developed \texttt{Q4RealBPP} with a focus on the new functionalities and features described in this study. In order for the problem to be solved, it must first be modeled as a CQM and then sent to the \texttt{LeapCQMHybrid} provided by D-Wave. This solver is part of D-Wave’s \textit{hybrid solver service} (HSS, \cite{HSS}), which is a portfolio of methods developed by D-Wave. These hybrid techniques balance classical and quantum computation for addressing optimization problems unable to be directly faced by purely quantum processors \cite{leapCQM}. 
At the time of this writing, the portfolio of hybrid methods available in HSS permits the solving of Binary Quadratic Models, Discrete Quadratic Models and, lastly, CQMs, such as the one formulated in this research. 

Having said this, the rest of this section is devoted to showcasing how \texttt{Q4RealBPP} deals with new realistic industrial contexts. These scenarios relate with \textit{heterogeneous bins}, \textit{item-bin associations} and \textit{delivery priorities}, respectively, using 3dBPP and 2dBPP instances. Also, we demonstrate the resolution of two different 1dBPP cases. Finally, for the sake of completeness, we test \texttt{Q4RealBPP} in two real-world oriented 3dBPP instances, considering not only the features described here but also further ones detailed in our previous work \cite{tecnalia2023hybrid}. Aiming to enhance the replicability of this work, the ten instances used and the results are openly available in \cite{BPPRep}. The format of these instances is based on the one provided in \cite{3dBPP}, and further described in \cite{osaba2023benchmark}.

\subsection{Heterogeneous Bins}\label{sec:hete}

\begin{figure}[!th]
    \centering
    \subfigure[Solution to \texttt{3dBPP\_het\_bins}.\label{fig:hete_a}]{\includegraphics[width=1.0\columnwidth]{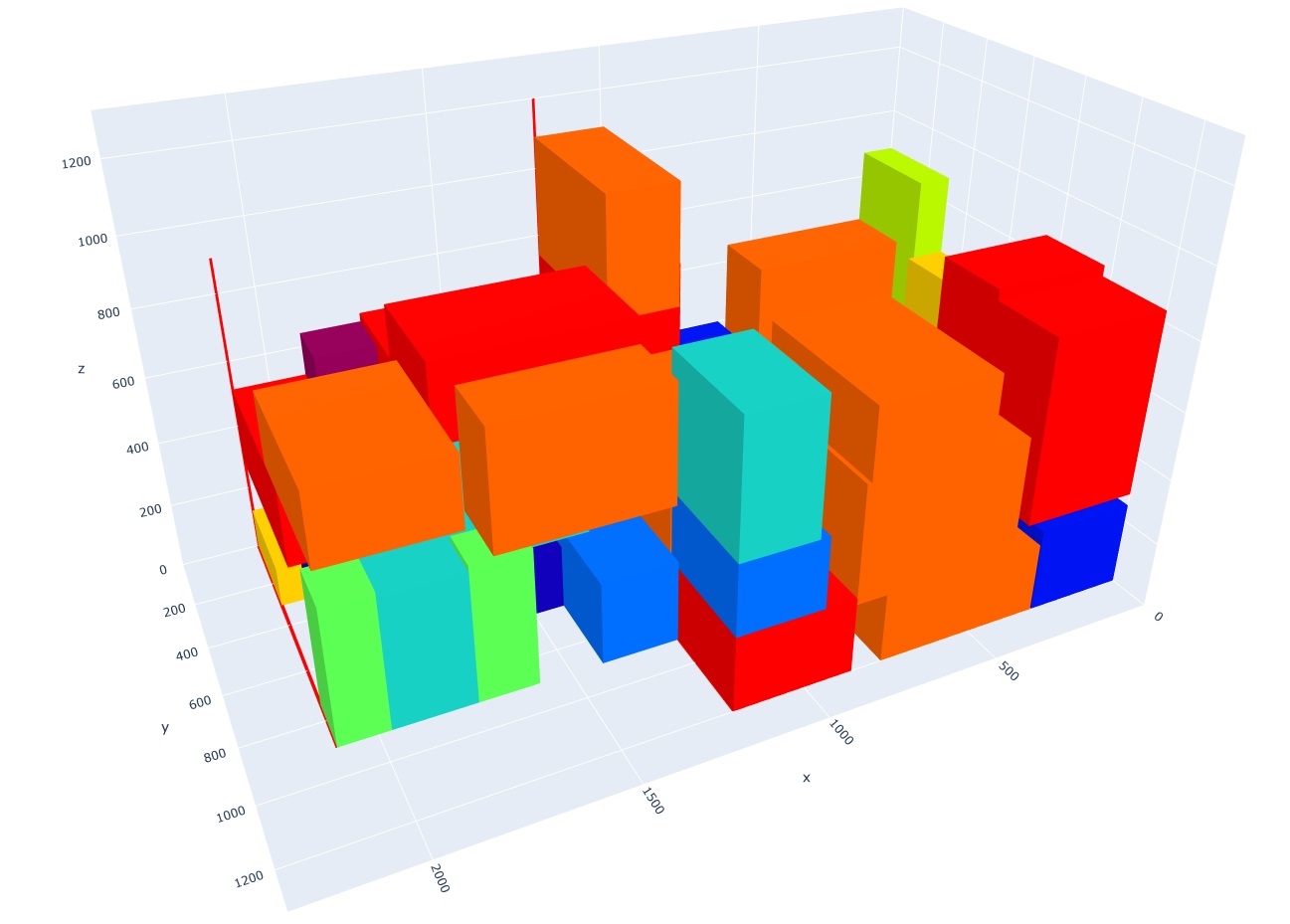}} \\
	\subfigure[Solution to \texttt{2dBPP\_het\_bins}.\label{fig:hete_b}]{\includegraphics[width=1.0\columnwidth]{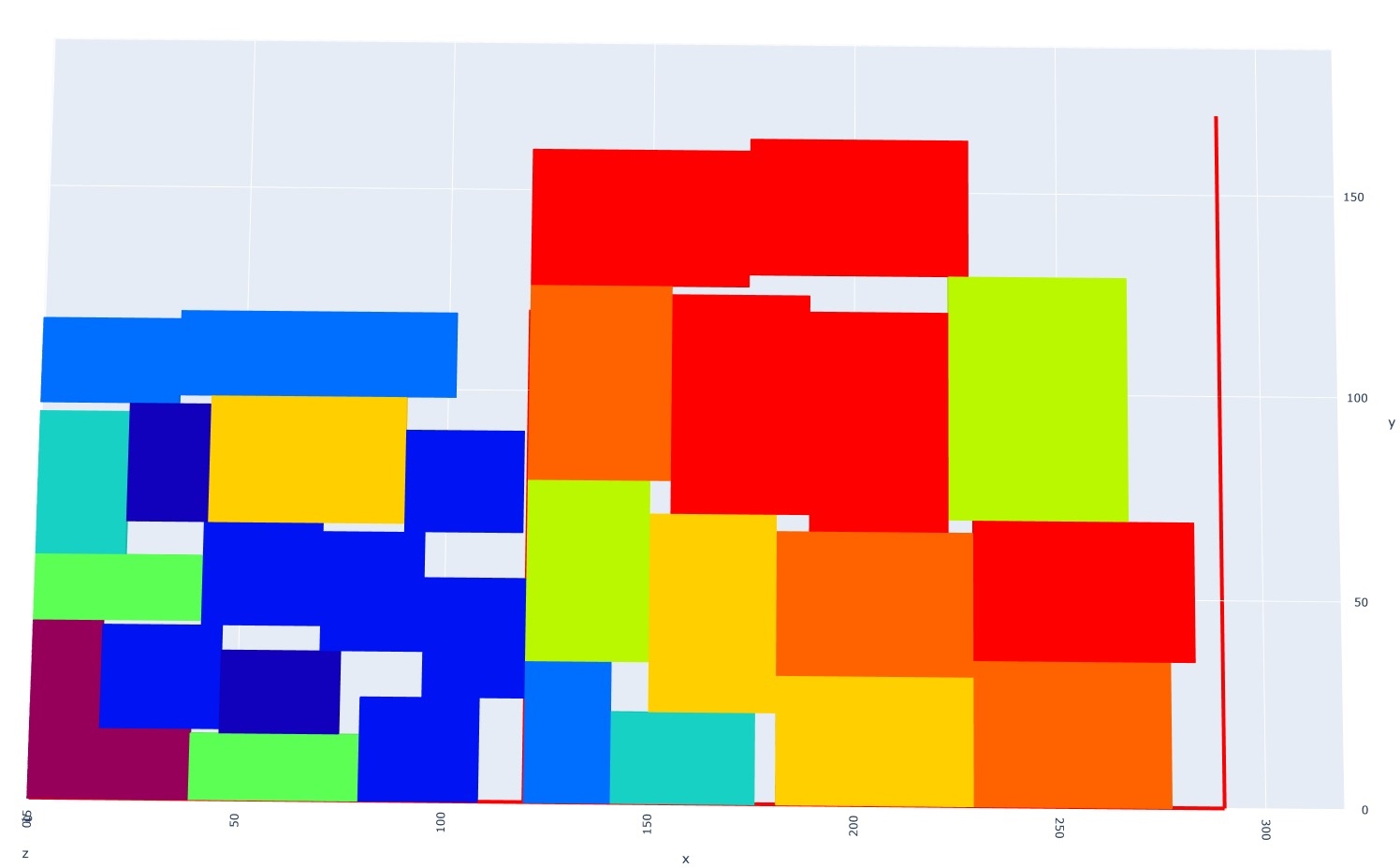}}\\
    \caption{Solutions provided by \texttt{Q4RealBPP} for the instances deeming heterogeneous bins.}
    \label{fig:Hete}
\end{figure}

On the one hand, we depict in Figure \ref{fig:hete_a} the result provided by \texttt{Q4RealBPP} for the instance coined \texttt{3dBPP\_het\_bins}. This instance is composed of $m$=51 items distributed among 10 categories and $n$=2 bins of $L_0$=$W_0$=$H_0$=1200 and $L_1$=$W_1$=$H_1$=900 dimensions. On the other hand, in Figure \ref{fig:hete_b} we represent the solution offered for the instance \texttt{2dBPP\_het\_bins}, with $m$=35 distributed among 10 categories and $n$=2 bins of $L_0$=$W_0$=120 and $L_1$=$W_1$=170 dimensions, respectively. Note that bin boundaries are represented by straight red lines in all figures shown in this manuscript. 

\subsection{Requirements for item-bin association}\label{sec:bin_req}

\begin{figure}[!t]
    \centering
    \subfigure[Solution to \texttt{3dBPP\_item\_bins}.]{\includegraphics[width=1.0\columnwidth]{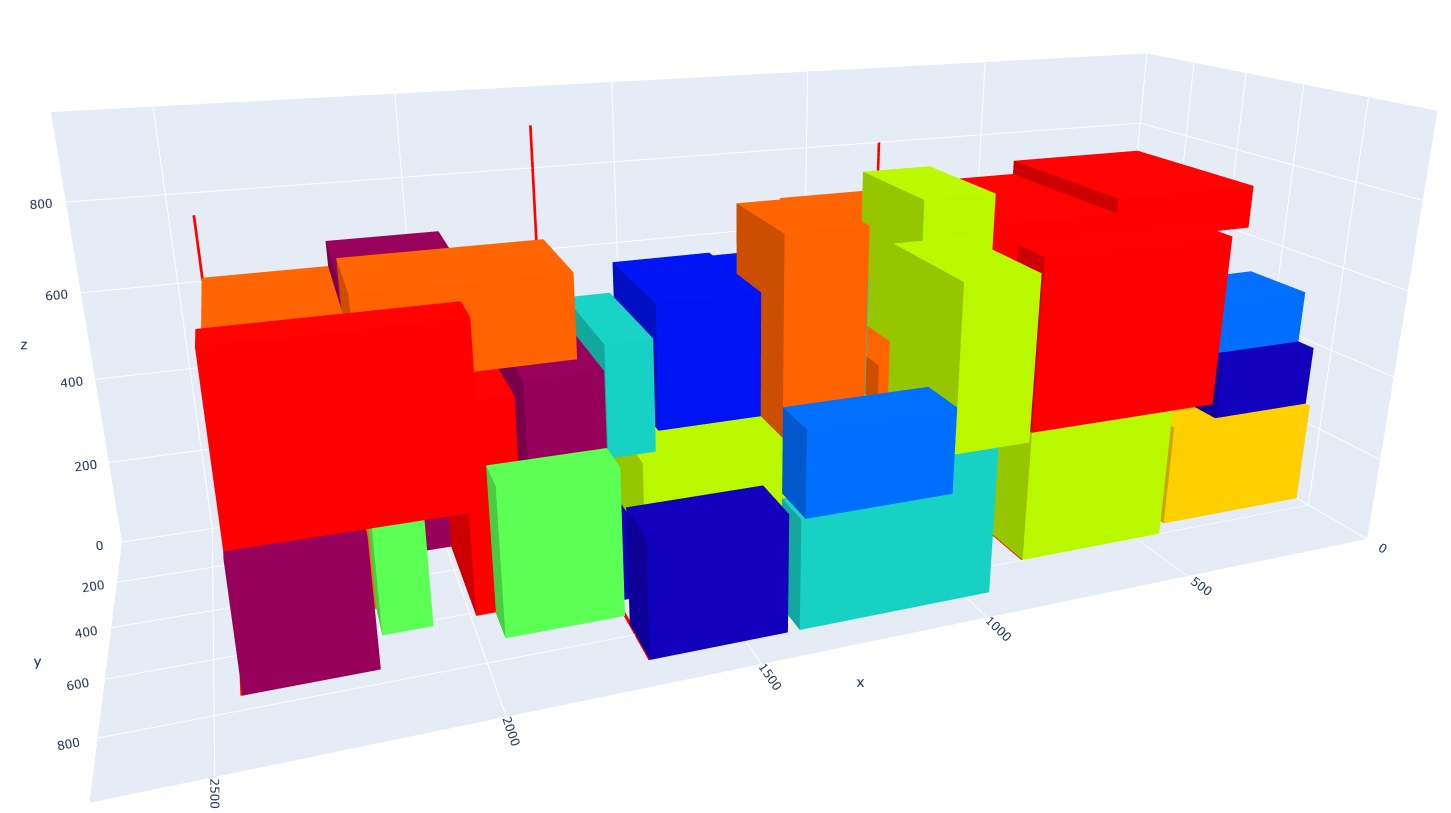}}\\
	\subfigure[Solution to \texttt{2dBPP\_item\_bins}.]{\includegraphics[width=1.0\columnwidth]{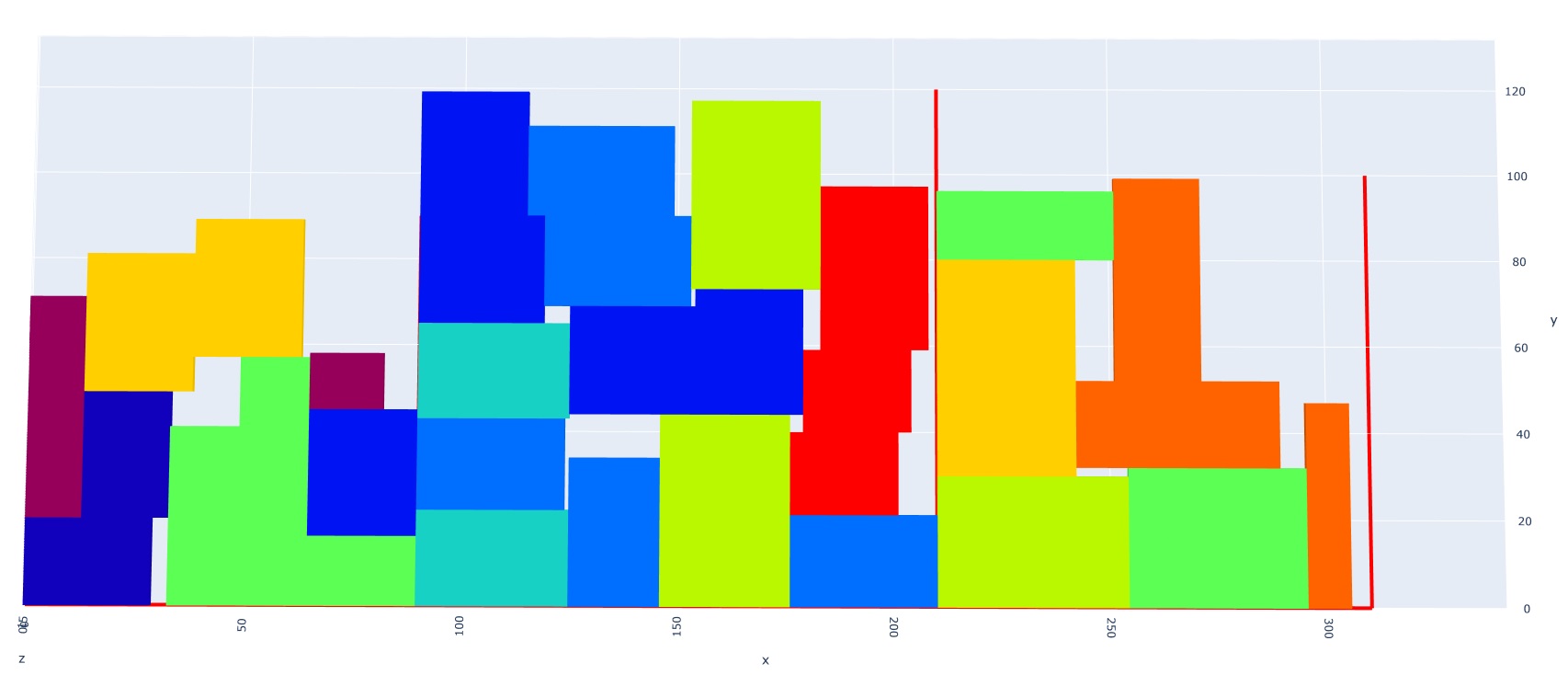}}\\
    \caption{Solutions provided by \texttt{Q4RealBPP} for the instances deeming requirements for item-bin associations. Bin \texttt{id}s are \{0,1,2\}, being \texttt{id}=0 the rightmost one, and \texttt{id}=2 the leftmost one.}
    \label{fig:binReqs}
\end{figure}

Table \ref{tab:bin_reqs} summarizes the requirements for item-bin association built for the two instances generated for this use case, named \texttt{3dBPP\_item\_bins} and \texttt{2dBPP\_item\_bins}. For each one, we represent the \texttt{id} of every item category and the bins allowed for each of these categories. Both instances count with $n$=3 bins, and $m$=57 items for the 3dBPP and $m$=40 for the 2dBPP. Solutions to these instances are shown in Figure \ref{fig:binReqs}.
\begin{table}[!b]
	\centering
	\caption{Summary of the item-bin associations for \texttt{3dBPP\_item\_bins} and \texttt{2dBPP\_item\_bins}. Each item category is represented by the number of items to store and the \texttt{ids} of the bins allowed. Both instances count with a pool of 3 bins with \texttt{ids} \{0,1,2\}.} \label{tab:bin_reqs}\vspace{-2mm}%
	\begin{tabularx}{\columnwidth}{YYYYYY}
		\toprule
		\multirow{2}*{Item ID} & \multirow{2}*{Color} & \multicolumn{2}{c}{3dBPP} & \multicolumn{2}{c}{2dBPP} \\ 
        & & Amount & Bins & Amount & Bins \\ \midrule
        0 & {\crule[Wine]} & 5 & 2 & 4 & 0 \\
        1 & {\crule[DarkBlue]}& 6 & 0,1,2 & 2 & 0 \\
        2 & {\crule[Blue]}& 3 & 0,1,2 & 5 & 0,1,2 \\
        3 & {\crule[LightBlue]}& 6 & 0,1,2 & 5 & 0,1,2 \\
        4 & {\crule[Teal]}& 8 & 1 & 2 & 0,1,2  \\
        5 & {\crule[Green]}& 6 & 0,2 & 6 & 0,2 \\
        6 & {\crule[Lime]}& 5 & 0,1 & 3 & 1,2 \\
        7 & {\crule[Yellow]}& 3 & 0 & 4 & 0,2 \\
        8 & {\crule[Orange]}& 7 & 1,2 & 5 & 2 \\
        9 & {\crule[Red]}& 8 & 0,2 & 4 & 1 \\ \bottomrule
	\end{tabularx}
\end{table}%

\subsection{Delivery Priorities}\label{sec:deliver}

\begin{figure}[!th]
    \centering
    \subfigure[Solution for \texttt{3dBPP\_del\_prior}.]{\includegraphics[width=0.85\columnwidth]{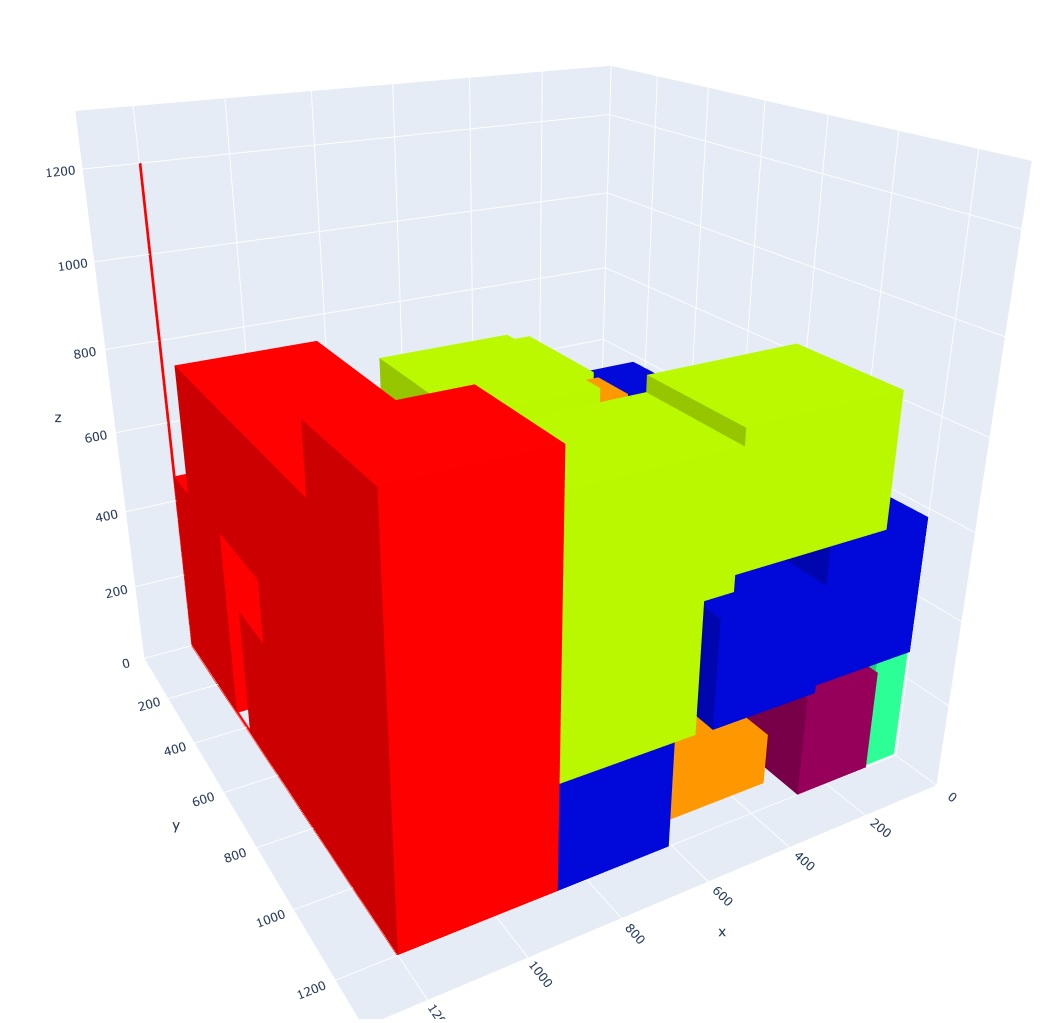}}\\
	\subfigure[Solution for \texttt{2dBPP\_del\_prior}.]{\includegraphics[width=0.8\columnwidth]{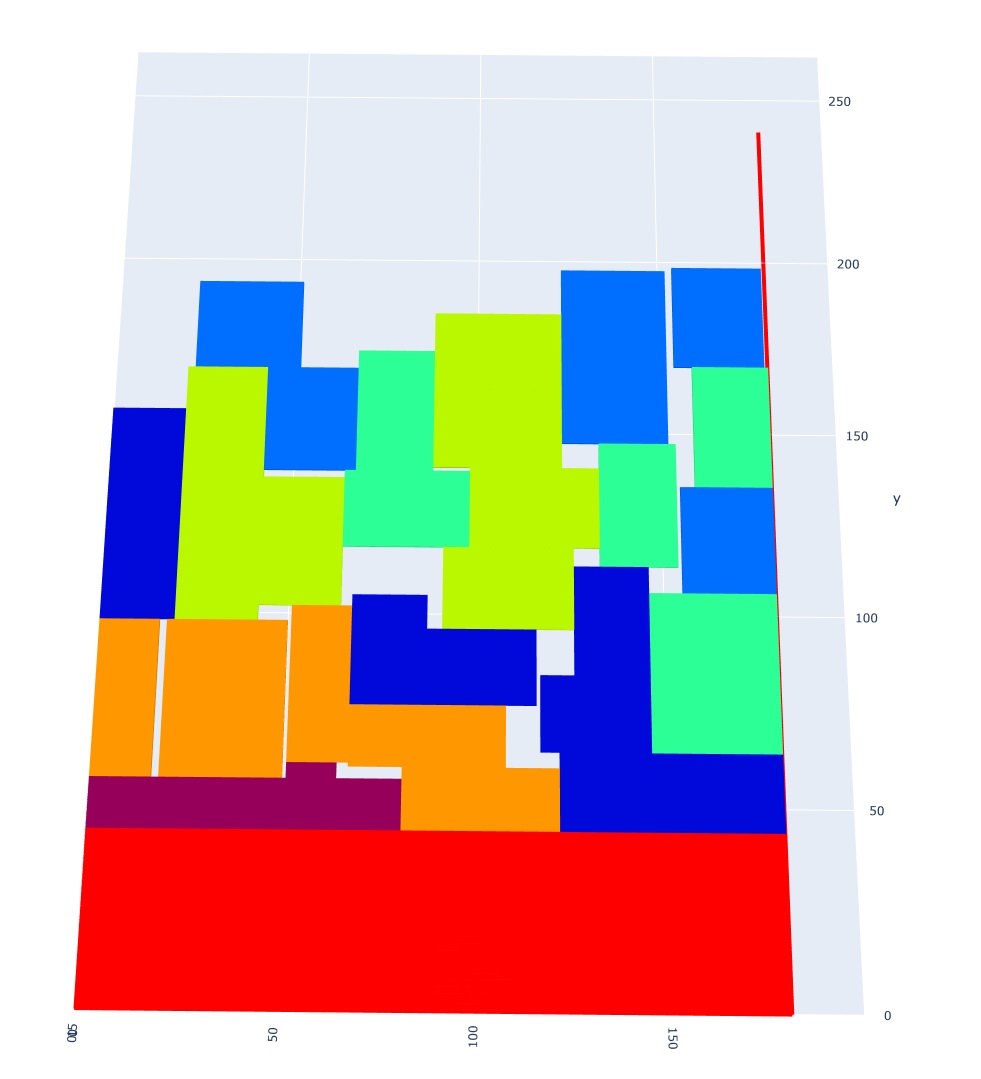}}\\
    \caption{Solutions provided by \texttt{Q4RealBPP} for the instances considering delivery priorities.}
    \label{fig:DelPrior}
\end{figure}

For both instances, named as \texttt{3dBPP\_del\_prior} and \texttt{2dBPP\_del\_prior}, the item category that should be prioritized is the \texttt{id}=9 (\crule[Red]), meaning that all packages of this class must be placed near an imaginary door for the 3dBPP ($y$ axis) and on the bottom of the bin for the 2dBPP ($x$ axis). In \texttt{3dBPP\_del\_prior} $m$ is equal to 42, while \texttt{2dBPP\_del\_prior} is composed of $m$=44 items. We represent the solutions given by \texttt{Q4RealBPP} for this use case in Figure \ref{fig:DelPrior}. 

\subsection{Solving 1dBPP instances}\label{sec:1d}

\begin{figure}[!th]
    \centering
    \subfigure[First bin in \texttt{1dBPP\_1} solution.]{\includegraphics[width=1.0\columnwidth]{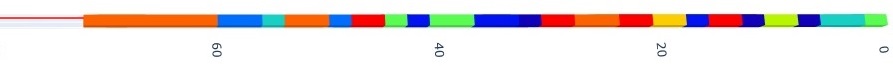}}\\
	\subfigure[Second bin in \texttt{1dBPP\_1} solution.]{\includegraphics[width=1.0\columnwidth]{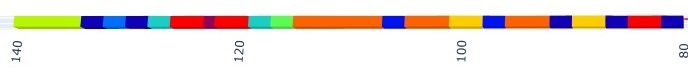}}\\
    \subfigure[First bin in \texttt{1dBPP\_2} solution.]{\includegraphics[width=1.0\columnwidth]{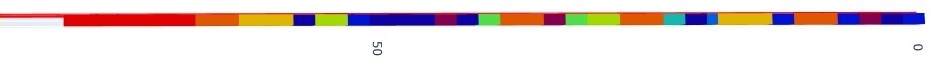}}\\
	\subfigure[Second bin in \texttt{1dBPP\_2} solution.]{\includegraphics[width=1.0\columnwidth]{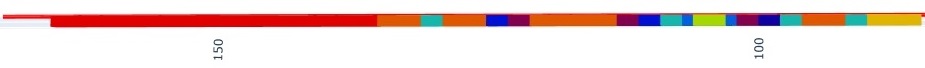}}\\
    \caption{Solutions provided by \texttt{Q4RealBPP} for \texttt{1dBPP\_1} (a \& b) and \texttt{1dBPP\_2} (c \& d) instances.}
    \label{fig:1dBPP}\vspace{-8mm}%
\end{figure}

Regarding the 1dBPP, we demonstrate the application of \texttt{Q4RealBPP} using two different instances. On the one hand, \texttt{1dBPP\_1} contains $m$=51 items, and $n$=2 bins, with $M_0$=80 and $M_1$=60. 
On the other hand, in \texttt{1dBPP\_2}, $m$=59, $n$=2 and $M_0$=$M_1$=85. 
We have also taken delivery priorities into account, storing items with \texttt{id}=9 (\crule[Red]) at the top of the bin. 
Figure \ref{fig:1dBPP} shows the solutions to both instances.

\subsection{Solving Real-World instances}\label{sec:rw}

Finally, and for the sake of completeness, we demonstrate in this subsection how \texttt{Q4RealBPP} can deal with two real-world oriented instances, which include not only the characteristics described in Section \ref{sec:Q4RealBPP} but also the ones shown in our previous work \cite{tecnalia2023hybrid}. Thus, the following two use cases have been built by combining multiple features developed in \texttt{Q4RealBPP}:
\begin{itemize}
    \item \texttt{3dBPP\_real\_world\_1}: $m$=41 distributed in 7 categories; $n$=1 with dimensions $L_0$=$W_0$=$H_0$=1600; a center of mass 
    $(\tilde{L},\tilde{W})$ = $(800,800)$; items with \texttt{id}=5 (\crule[Orange]) are so heavy that must not be placed beneath any package of other category; and delivery priority for items with \texttt{id}=6 (\crule[Red]).
    \item \texttt{3dBPP\_real\_world\_2}: $m$=55 items distributed among 10 categories; $n$=3 with $L_0$=$W_0$=$H_0$=750, $L_1$=$W_1$=$H_1$=800 and $L_2$=$W_2$=$H_2$=900; items with the \texttt{id}=1 (\crule[DarkBlue]) must not be stored in the same bin as \texttt{id}=2 (\crule[Blue]) and \texttt{id}=3 (\crule[LightBlue]); item-bin associations equal to what detailed in Table \ref{tab:bin_reqs} for the 3dBPP instance; items with \texttt{id}=9 (\crule[Red]) are so heavy that they must not be placed beneath any package of another category; and delivery priority for items with \texttt{id}=8 (\crule[Orange]).
\end{itemize}%
\begin{figure}[!t]
    \centering
    \subfigure[Solution for \texttt{3dBPP\_real\_world\_1}.]{\includegraphics[width=0.8\columnwidth]{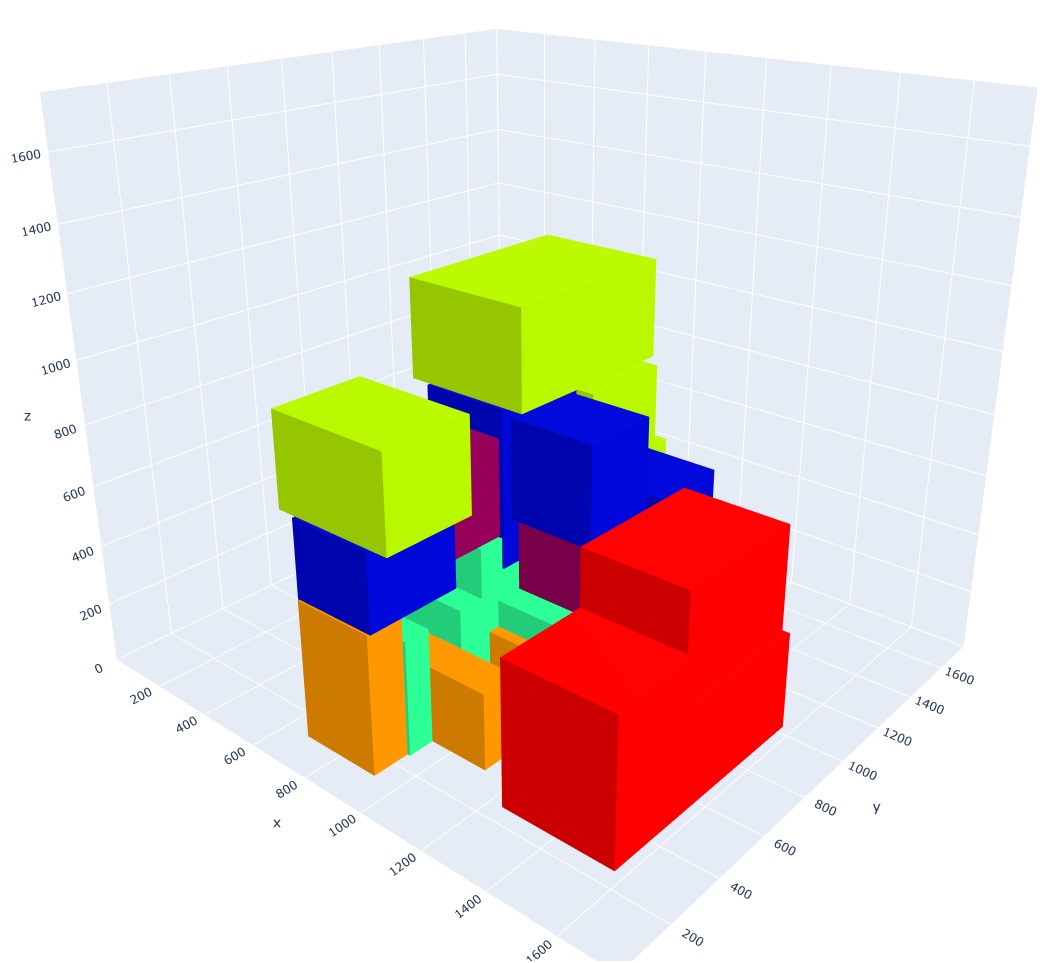}}\hspace{2mm}%
	\subfigure[Solution for \texttt{3dBPP\_real\_world\_2}.]{\includegraphics[width=0.9\columnwidth]{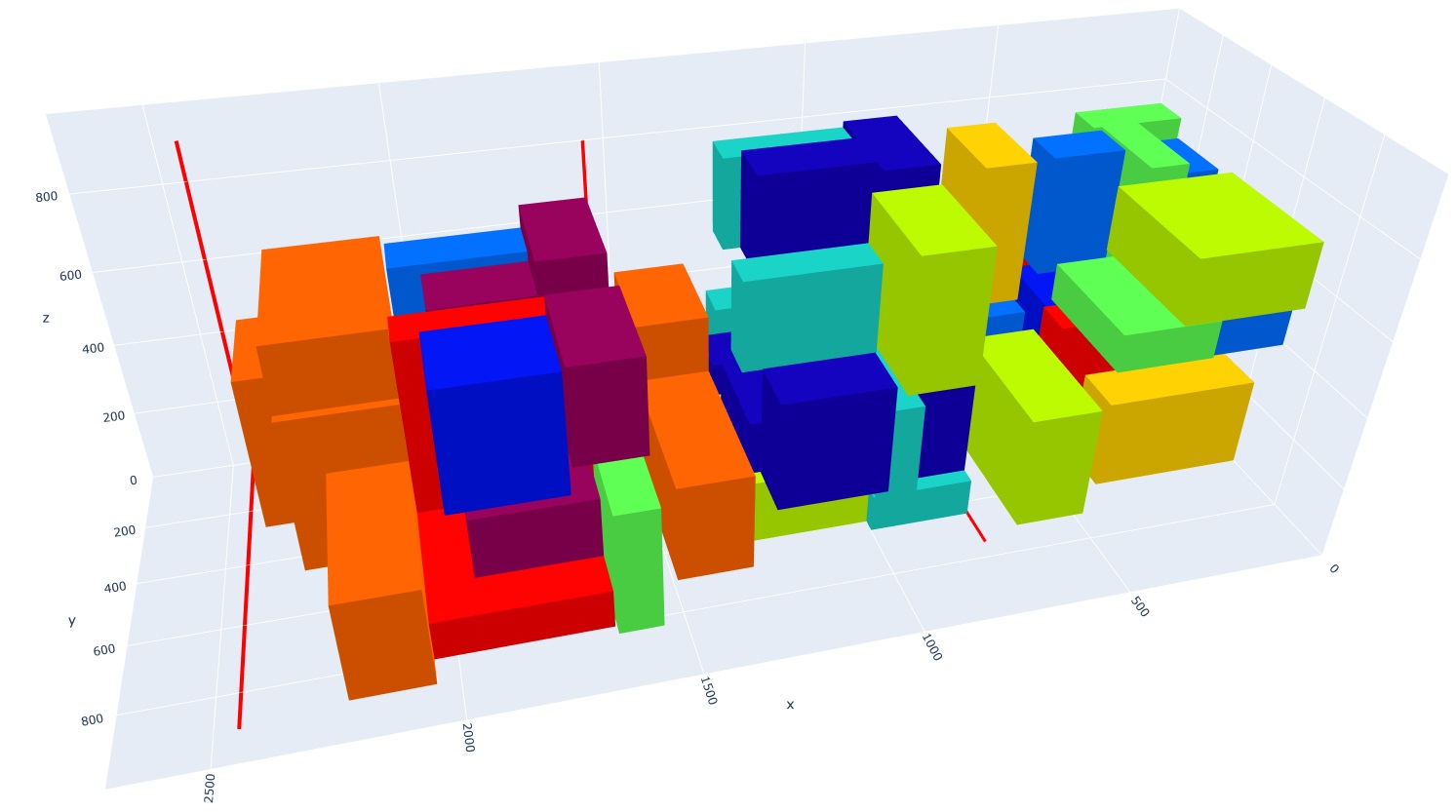}}\hspace{2mm}%
    \caption{Solutions provided by \texttt{Q4RealBPP} for the real-world oriented instances. Bin \texttt{id}s are \{0,1,2\}, being \texttt{id}=0 the rightmost one, and \texttt{id}=2 the leftmost one.}
    \label{fig:RealWorld}
\end{figure}

\section{Conclusions and further work}\label{sec:conc}

This paper has enriched our previous work \cite{tecnalia2023hybrid} with new industrial applications. Thanks to the work detailed in this manuscript, our tool is able to deal with more realistic and logistics oriented BPP scenarios. More specifically, we have described and tested four different features: \textit{i}) the generalization to 1dBPP and 2dBPP, \textit{ii}) the possibility of having heterogeneous bins, \textit{iii}) requirements for item-bin associations, and \textit{iv}) the introduction of delivery priorities. We have planned for future research to develop additional functionalities, such as deeming two centers of mass in the same bin or the consideration of constraints as soft restrictions, and test additional realistic scenarios, such as ones considering delivery priorities in both $y$ and $z$ axes. Also, we will work on improving the overall quality of the solutions provided by \texttt{Q4RealBPP}. Finally, we also contemplate as further work the introduction of advanced paradigms such as transfer optimization.

\bibliographystyle{ieeeconf}

\begin{thebibliography}{10}
\providecommand{\url}[1]{#1}
\csname url@samestyle\endcsname
\providecommand{\newblock}{\relax}
\providecommand{\bibinfo}[2]{#2}
\providecommand{\BIBentrySTDinterwordspacing}{\spaceskip=0pt\relax}
\providecommand{\BIBentryALTinterwordstretchfactor}{4}
\providecommand{\BIBentryALTinterwordspacing}{\spaceskip=\fontdimen2\font plus
\BIBentryALTinterwordstretchfactor\fontdimen3\font minus
  \fontdimen4\font\relax}
\providecommand{\BIBforeignlanguage}[2]{{%
\expandafter\ifx\csname l@#1\endcsname\relax
\typeout{** WARNING: IEEEtran.bst: No hyphenation pattern has been}%
\typeout{** loaded for the language `#1'. Using the pattern for}%
\typeout{** the default language instead.}%
\else
\language=\csname l@#1\endcsname
\fi
#2}}
\providecommand{\BIBdecl}{\relax}
\BIBdecl

\bibitem{bruton2016packing}
J.~T. Bruton, T.~G. Nelson, T.~K. Zimmerman, J.~D. Fernelius, S.~P. Magleby,
  and L.~L. Howell, ``Packing and deploying soft origami to and from
  cylindrical volumes with application to automotive airbags,'' \emph{Royal
  Society open science}, vol.~3, no.~9, p. 160429, 2016.

\bibitem{maddaloni2016bin}
A.~Maddaloni, V.~Colla, G.~Nastasi, M.~Del~Seppia, and V.~Iannino, ``A bin
  packing algorithm for steel production,'' in \emph{2016 European Modelling
  Symposium (EMS)}.\hskip 1em plus 0.5em minus 0.4em\relax IEEE, 2016, pp.
  19--24.

\bibitem{kim2020online}
G.~Kim and I.~Moon, ``Online banner advertisement scheduling for advertising
  effectiveness,'' \emph{Computers \& Industrial Engineering}, vol. 140, p.
  106226, 2020.

\bibitem{tian2023learning}
R.~Tian, C.~Kang, J.~Bi, Z.~Ma, Y.~Liu, S.~Yang, and F.~Li, ``Learning to
  multi-vehicle cooperative bin packing problem via sequence-to-sequence policy
  network with deep reinforcement learning model,'' \emph{Computers \&
  Industrial Engineering}, p. 108998, 2023.

\bibitem{gzara2020pallet}
F.~Gzara, S.~Elhedhli, and B.~C. Yildiz, ``The pallet loading problem:
  Three-dimensional bin packing with practical constraints,'' \emph{European
  Journal of Operational Research}, vol. 287, no.~3, pp. 1062--1074, 2020.

\bibitem{you2014design}
S.~You and S.~Ji, ``Design of a multi-robot bin packing system in an automatic
  warehouse,'' in \emph{2014 11th International Conference on Informatics in
  Control, Automation and Robotics (ICINCO)}, vol.~2.\hskip 1em plus 0.5em
  minus 0.4em\relax IEEE, 2014, pp. 533--538.

\bibitem{zhao2021mathematical}
P.~Zhao, H.~Guan, H.~Wei, and S.~Liu, ``Mathematical modeling and heuristic
  approaches to optimize shared parking resources: A case study of beijing,
  china,'' \emph{Transportation Research Interdisciplinary Perspectives},
  vol.~9, p. 100317, 2021.

\bibitem{gyongyosi2019survey}
L.~Gyongyosi and S.~Imre, ``A survey on quantum computing technology,''
  \emph{Computer Science Review}, vol.~31, pp. 51--71, 2019.

\bibitem{wang2021shaping}
S.~Wang, Z.~Pei, C.~Wang, and J.~Wu, ``Shaping the future of the application of
  quantum computing in intelligent transportation system,'' \emph{Intelligent
  and Converged Networks}, vol.~2, no.~4, pp. 259--276, 2021.

\bibitem{osaba2022systematic}
E.~Osaba, E.~Villar-Rodriguez, and I.~Oregi, ``A systematic literature review
  of quantum computing for routing problems,'' \emph{IEEE Access}, 2022.

\bibitem{boyapati2020quantum}
S.~Boyapati, S.~R. Swarna, and A.~Kumar, ``Quantum neural networks for dynamic
  route identification to avoid traffic,'' in \emph{2020 Fourth International
  Conference on I-SMAC (IoT in Social, Mobile, Analytics and
  Cloud)(I-SMAC)}.\hskip 1em plus 0.5em minus 0.4em\relax IEEE, 2020, pp.
  1018--1022.

\bibitem{tecnalia2023hybrid}
S.~V.~Romero, E.~Osaba, E.~Villar-Rodriguez, I.~Oregi, and Y.~Ban, ``Hybrid
  approach for solving real-world bin packing problem instances using quantum
  annealers,'' \emph{Scientific Reports}, vol.~13, no.~1, p. 11777, 2023.

\bibitem{martello2000three}
S.~Martello, D.~Pisinger, and D.~Vigo, ``The three-dimensional bin packing
  problem,'' \emph{Operations research}, vol.~48, no.~2, pp. 256--267, 2000.

\bibitem{munien2021metaheuristic}
C.~Munien and A.~E. Ezugwu, ``Metaheuristic algorithms for one-dimensional
  bin-packing problems: A survey of recent advances and applications,''
  \emph{Journal of Intelligent Systems}, vol.~30, no.~1, pp. 636--663, 2021.

\bibitem{lodi2002recent}
A.~Lodi, S.~Martello, and D.~Vigo, ``Recent advances on two-dimensional bin
  packing problems,'' \emph{Discrete Applied Mathematics}, vol. 123, no. 1-3,
  pp. 379--396, 2002.

\bibitem{hu2020greedy}
X.~Hu, J.~Li, and J.~Cui, ``Greedy adaptive search: A new approach for
  large-scale irregular packing problems in the fabric industry,'' \emph{IEEE
  Access}, vol.~8, pp. 91\,476--91\,487, 2020.

\bibitem{leung2011extended}
S.~C. Leung, X.~Zhou, D.~Zhang, and J.~Zheng, ``Extended guided tabu search and
  a new packing algorithm for the two-dimensional loading vehicle routing
  problem,'' \emph{Computers \& Operations Research}, vol.~38, no.~1, pp.
  205--215, 2011, project Management and Scheduling.

\bibitem{lodi2002heuristic}
A.~Lodi, S.~Martello, and D.~Vigo, ``Heuristic algorithms for the
  three-dimensional bin packing problem,'' \emph{European Journal of
  Operational Research}, vol. 141, no.~2, pp. 410--420, 2002.

\bibitem{paquay2018mip}
C.~Paquay, S.~Limbourg, M.~Schyns, and J.~F. Oliveira, ``Mip-based constructive
  heuristics for the three-dimensional bin packing problem with transportation
  constraints,'' \emph{International Journal of Production Research}, vol.~56,
  no.~4, pp. 1581--1592, 2018.

\bibitem{parreno2010hybrid}
F.~Parre{\~n}o, R.~Alvarez-Vald{\'e}s, J.~F. Oliveira, and J.~M. Tamarit, ``A
  hybrid grasp/vnd algorithm for two-and three-dimensional bin packing,''
  \emph{Annals of Operations Research}, vol. 179, no.~1, pp. 203--220, 2010.

\bibitem{do2021practical}
O.~X. do~Nascimento, T.~A. de~Queiroz, and L.~Junqueira, ``Practical
  constraints in the container loading problem: Comprehensive formulations and
  exact algorithm,'' \emph{Computers \& Operations Research}, vol. 128, p.
  105186, 2021.

\bibitem{silva2019exact}
E.~F. Silva, T.~Wauters \emph{et~al.}, ``Exact methods for three-dimensional
  cutting and packing: A comparative study concerning single container
  problems,'' \emph{Computers \& Operations Research}, vol. 109, pp. 12--27,
  2019.

\bibitem{jiang2012hybrid}
J.~Jiang and L.~Cao, ``A hybrid simulated annealing algorithm for
  three-dimensional multi-bin packing problems,'' in \emph{2012 International
  Conference on Systems and Informatics (ICSAI2012)}.\hskip 1em plus 0.5em
  minus 0.4em\relax IEEE, 2012, pp. 1078--1082.

\bibitem{weiss2021solving}
M.~Weiss~Cohen, ``Solving a profited 3d bin packing problem using a hybrid
  genetic algorithm,'' in \emph{ASME International Mechanical Engineering
  Congress and Exposition}, vol. 85604.\hskip 1em plus 0.5em minus 0.4em\relax
  American Society of Mechanical Engineers, 2021, p. V006T06A001.

\bibitem{hu2017solving}
H.~Hu, X.~Zhang, X.~Yan, L.~Wang, and Y.~Xu, ``Solving a new 3d bin packing
  problem with deep reinforcement learning method,'' \emph{arXiv preprint
  arXiv:1708.05930}, 2017.

\bibitem{seskir2023democratization}
Z.~C. Seskir, S.~Umbrello, C.~Coenen, and P.~E. Vermaas, ``Democratization of
  quantum technologies,'' \emph{Quantum Science and Technology}, vol.~8, no.~2,
  p. 024005, 2023.

\bibitem{harwood2021formulating}
S.~Harwood, C.~Gambella, D.~Trenev, A.~Simonetto, D.~Bernal, and D.~Greenberg,
  ``Formulating and solving routing problems on quantum computers,'' \emph{IEEE
  Transactions on Quantum Engineering}, vol.~2, pp. 1--17, 2021.

\bibitem{ajagekar2019quantum}
A.~Ajagekar and F.~You, ``Quantum computing for energy systems optimization:
  Challenges and opportunities,'' \emph{Energy}, vol. 179, pp. 76--89, 2019.

\bibitem{orus2019quantum}
R.~Or{\'u}s, S.~Mugel, and E.~Lizaso, ``Quantum computing for finance: Overview
  and prospects,'' \emph{Reviews in Physics}, vol.~4, p. 100028, 2019.

\bibitem{de2022hybrid}
M.~G. de~Andoin, E.~Osaba, I.~Oregi, E.~Villar-Rodriguez, and M.~Sanz, ``Hybrid
  quantum-classical heuristic for the bin packing problem,'' in
  \emph{Proceedings of the Genetic and Evolutionary Computation Conference
  Companion}, 2022, pp. 2214--2222.

\bibitem{garcia2022comparative}
M.~G. De~Andoin, I.~Oregi, E.~Villar-Rodriguez, E.~Osaba, and M.~Sanz,
  ``Comparative benchmark of a quantum algorithm for the bin packing problem,''
  in \emph{2022 IEEE Symposium Series on Computational Intelligence (SSCI)},
  2022, pp. 930--937.

\bibitem{bozhedarov2023quantum}
A.~Bozhedarov, A.~Boev, S.~Usmanov, G.~Salahov, E.~Kiktenko, and A.~Fedorov,
  ``Quantum and quantum-inspired optimization for solving the minimum bin
  packing problem,'' \emph{arXiv preprint arXiv:2301.11265}, 2023.

\bibitem{3dBPP}
\BIBentryALTinterwordspacing
{D-Wave Ocean Developers Team}, ``{\texttt{3d-bin-packing} (GitHub
  repository)},'' 6 2022, {Last retrieved \today}. [Online]. Available:
  \url{https://github.com/dwave-examples/3d-bin-packing}
\BIBentrySTDinterwordspacing

\bibitem{osaba2023benchmark}
E.~Osaba, E.~Villar-Rodriguez, and S.~V.~Romero, ``Benchmark dataset and
  instance generator for real-world three-dimensional bin packing problems,''
  \emph{Data in Brief}, p. 109309, 2023.

\bibitem{leapCQM}
{D-Wave Developers}, ``{Measuring Performance of the Leap Constrained Quadratic
  Model Solver},'' D-Wave Systems Inc., Tech. Rep. 14-1065A-A, 11 2022.

\bibitem{anand2020bin}
S.~Anand and S.~Guericke, ``A bin packing problem with mixing constraints for
  containerizing items for logistics service providers,'' in
  \emph{Computational Logistics: 11th International Conference, ICCL 2020,
  Enschede, The Netherlands, September 28--30, 2020, Proceedings}.\hskip 1em
  plus 0.5em minus 0.4em\relax Springer, 2020, pp. 342--355.

\bibitem{Preskill2018}
J.~Preskill, ``Quantum {C}omputing in the {NISQ} era and beyond,''
  \emph{{Quantum}}, vol.~2, p.~79, Aug. 2018.

\bibitem{cid2020positions}
N.~M. Cid-Garcia and Y.~A. Rios-Solis, ``Positions and covering: A two-stage
  methodology to obtain optimal solutions for the 2d-bin packing problem,''
  \emph{PLOS ONE}, vol.~15, no.~4, p. e0229358, Apr 2020.

\bibitem{HSS}
{D-Wave Developers}, ``{D-Wave Hybrid Solver Service: An Overview},'' D-Wave
  Systems Inc., Tech. Rep. 14-1039A-B, 05 2020.

\bibitem{BPPRep}
E.~Osaba, S.~V.~Romero, and E.~Villar-Rodriguez, ``Benchmark dataset for
  logistic-oriented bin packing problems,''
  \url{http://dx.doi.org/10.17632/9ts4rvkc5s.1}, 2023, online at Mendeley Data.

\end{thebibliography}

\end{document}